\documentclass[10pt,twocolumn,letterpaper]{article}

\usepackage{cvpr}              

\usepackage[table,xcdraw]{xcolor} 
\usepackage{multirow}
\usepackage[most]{tcolorbox}   
\usepackage{listings}          
\usepackage{graphicx}          
\usepackage{amssymb}           

\definecolor{sys-bg}{RGB}{240, 240, 245}   
\definecolor{bot-bg}{RGB}{255, 250, 240}   
\definecolor{code-bg}{RGB}{248, 248, 248}  

\newtcolorbox{dialoguebox}[2][]{
    colback=#2,
    colframe=gray!40,
    boxrule=0.5pt,
    arc=2pt,
    left=2pt, right=2pt, top=2pt, bottom=2pt,
    fonttitle=\bfseries\scriptsize,
    fontupper=\scriptsize, 
    title=#1,
    enhanced
}

\newtcblisting{promptbox}{
  listing only,
  breakable,
  enhanced,
  colback=gray!10,
  colframe=gray!50,
  boxrule=0.5pt,
  arc=3pt,
  left=6pt, right=6pt, top=6pt, bottom=6pt,
  listing options={
    basicstyle=\ttfamily\scriptsize\linespread{0.9},
    breaklines=true,
    columns=fullflexible,
    showstringspaces=false,
    aboveskip=0pt,
    belowskip=0pt
  }
}

\lstdefinelanguage{json}{
    basicstyle=\normalfont\ttfamily,
    string=[s]{"}{"},
    stringstyle=\color{cvprblue}, 
    comment=[l]{:},
    commentstyle=\color{black},
}

\usepackage{amsmath,amssymb,amsfonts}
\usepackage{dsfont}     
\usepackage{enumitem}   
\usepackage{bbm}
\usepackage[ruled,vlined,linesnumbered,noend]{algorithm2e} 
\DontPrintSemicolon
\SetAlFnt{\small}\SetAlCapFnt{\small}\SetAlCapNameFnt{\small}
\newcommand{\E}{\mathbb{E}}
\newcommand{\KL}{\mathrm{KL}}

\definecolor{cvprblue}{rgb}{0.21,0.49,0.74}
\usepackage[pagebackref,breaklinks,colorlinks,allcolors=cvprblue]{hyperref}

\def\paperID{} 
\def\confName{CVPR}
\def\confYear{2026}
\title{EVA: Efficient Reinforcement Learning for End-to-End Video Agent}
\author{
Yaolun Zhang\thanks{Equal contribution \quad $^\ddagger$Project Lead \quad $^\dagger$Corresponding author: {\tt\small luotto@sensetime.com} \quad  Our Code and model are at \href{https://github.com/wangruohui/EfficientVideoAgent}{this link}.} \quad
Ruohui Wang$^*$ \quad
Jiahao Wang$^{*,\ddagger}$ \quad
Yepeng Tang \quad
Xuanyu Zheng \\
Haonan Duan \quad
Hao Lu \quad
Hanming Deng \quad
Lewei Lu$^\dagger$ \\
\\
SenseTime Research \\
}

\begin{document}
\maketitle
\begin{abstract}
Video understanding with multimodal large language models (MLLMs) remains challenging due to the long token sequences of videos, which contain extensive temporal dependencies and redundant frames. Existing approaches typically treat MLLMs as passive recognizers, processing entire videos or uniformly sampled frames without adaptive reasoning. Recent agent-based methods introduce external tools, yet still depend on manually designed workflows and perception-first strategies, resulting in inefficiency on long videos. We present EVA, an Efficient Reinforcement Learning framework for End-to-End Video Agent, which enables planning-before-perception through iterative summary–plan–action–reflection reasoning. EVA autonomously decides what to watch, when to watch, and how to watch, achieving query-driven and efficient video understanding. To train such agents, we design a simple yet effective three-stage learning pipeline—comprising supervised fine-tuning (SFT), Kahneman–Tversky Optimization (KTO), and Group Relative Policy Optimization (GRPO)—that bridges supervised imitation and reinforcement learning. We further construct high-quality datasets for each stage, supporting stable and reproducible training. We evaluate EVA on six video understanding benchmarks, demonstrating its comprehensive capabilities. Compared with existing baselines, EVA achieves a substantial improvement of 6--12\% over general MLLM baselines and a further 1--3\% gain over prior adaptive agent methods. 
\end{abstract}    
\section{Introduction}
\label{sec:intro}



\begin{figure*}
    
    \centering
    \includegraphics[width=1\linewidth]{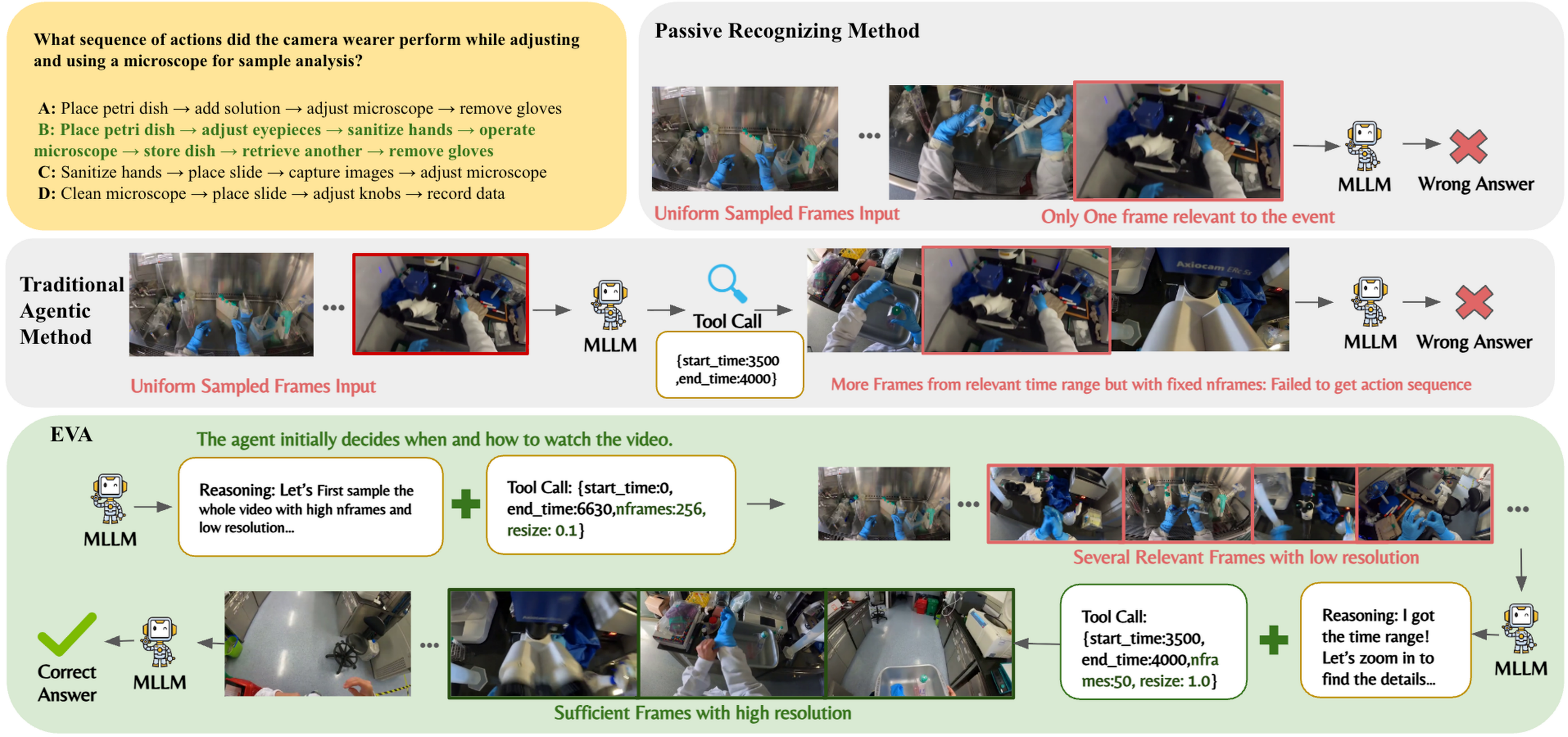}
    \caption{Given a question that requires the MLLM to figure out action sequences from an extremely long video (over 6600 seconds), the traditional uniformed sampling method is limited by the content length of the MLLM, and it is extremely hard for it to sample all the keyframes to answer the question correctly. As for the Traditional Agentic Method, the agent will also be given the uniformly sampled frames along with the video, which already occupy a lot of context. Although the agent can call tools to extract frames from a specific time range, the tool is rigid and the agent cannot adjust the fps and resolution, which leads to potential information loss. However, in EVA, the agent can arrange the tokens wisely. It can first watch the whole video with low resolution and high fps to get an overview of the video without costing too many visual tokens. After it finds the key time range, it will extract frames with high fps and high resolution, which leads to the correct answer.}
    \label{fig:fig1}
\end{figure*}
Video understanding has emerged as a cornerstone of multimodal intelligence~\citep{bai2025qwen25vltechnicalreport,zhang2025llavavideovideoinstructiontuning}, enabling a wide range of applications such as video question answering, retrieval, and embodied perception. 
As multimodal large language models (MLLMs) become increasingly capable of integrating vision, language, and reasoning~\citep{feng2025videor1reinforcingvideoreasoning}, a new research frontier is opening up---transforming them from passive perception models into active agents. 
This naturally raises a fundamental question: \textit{How can an MLLM-based Agent decide when and how to watch a video autonomously?}


Most existing video understanding systems still treat MLLMs as passive recognizers—they process entire videos or uniformly sampled frames to generate responses, without any notion of selective attention or adaptive reasoning~\citep{zhang2025llavavideovideoinstructiontuning, tang2025adaptive, hu2025m}, as illustrated in Figure~\ref{fig:fig1}.
Recent agent-based approaches take a step forward by introducing external tools such as frame-selection modules~\citep{he2025framethinker, pan2025timesearch, wang2025video}.
However, these pipelines remain largely handcrafted—built upon fixed parameters, rigid workflows, and limited exploration capabilities (e.g., fixed sampling rates).
Moreover, even these “agent-based” methods typically start their reasoning after being fed a set of uniformly sampled frames together with the textual query, making them still perception-first rather than truly planning-driven, which results in redundant visual processing and limited reasoning efficiency on long videos.

To bridge this gap, we advocate a \textbf{planning-before-perception} paradigm, where the agent first reasons solely from the textual query to decide what to watch, when to watch, and how to watch, before engaging with any visual input. We formulate such video understanding as an iterative process of \textit{summary–planning–action–reflection}. This paradigm allows the agent to progressively refine its perception and reasoning in response to the query, selectively attending to informative moments while avoiding unnecessary computation. Through this lens, an MLLM evolves from a passive video recognizer into an active, adaptive, and autonomous agentic watcher.

At the heart of our approach lies an \textbf{iterative perception, reasoning, and tool usage paradigm} that couples visual summary, planning, tool calling, and reflection thinking. 
The central challenge is to enable the model to operate effectively within this reasoning loop: learning how to generate the initial tool call based solely on the query without watching the video, how to continue reasoning when the available visual information is insufficient, and how to avoid over-exploration or becoming trapped in unnecessary iterations.

To address this challenge, we introduce a three-stage training strategy. In the initial training stage, we construct a \textbf{Supervised Fine-Tuning (SFT) Cold-Start} dataset to instill core video-agent capabilities: tool-call formatting, interleaved image–text reasoning, frame-level understanding, and basic frame-selecting strategies. This cold-start phase supplies the model with a stable behavioral prior that eases later, more aggressive optimization.
The second stage uses a \textbf{Kahneman–Tversky Optimization (KTO)}~\citep{ethayarajh2024ktomodelalignmentprospect} dataset composed of both successful and failed strategy trajectories. KTO guides the agent to prefer effective strategies while avoiding common failure modes; by correcting these known bad cases prior to GRPO~\citep{shao2024deepseekmathpushinglimitsmathematical}, it improves convergence, robustness, and stability during online policy optimization.
Third, we introduce an online reinforcement learning stage based on \textbf{Group Relative Policy Optimization (GRPO)}, which employs multiple data-driven rewards for both open-ended and multiple-choice question answering (QA).
These standard yet flexible reward signals balance reasoning depth with computational efficiency, enabling scalable and adaptive policy learning for video understanding.

Together, these mechanisms enable the agent to learn an adaptive policy for multi-round perception, planning, and tool usage, ensuring effective video understanding while controlling redundant computation.

To ensure stable and reproducible agentic reinforcement learning, we curate and construct a series of high-quality datasets for both the SFT cold-start and reinforcement learning stages. Specifically, we introduce the EVA-SFT, EVA-KTO, and EVA-RL datasets.
The EVA-SFT comprises 10k high-quality samples covering both general and task-specific agent training data.
The EVA-KTO includes 11k labeled frame-selection strategies, capturing diverse success and failure trajectories to guide strategy optimization.
The EVA-RL contains 9.6k open-ended video QA pairs and 1.1k multiple choice questions. 


Overall, our main contributions are summarized as follows:

\begin{itemize}
    \item \textbf{A novel and efficient RL-based video agent (EVA).} We propose a planning-before-perception framework featuring iterative summary–plan–action–reflection cycles, enabling efficient and interpretable video understanding.
    \item \textbf{Simple yet effective three-stage end to end training pipeline.} Our framework combines SFT cold-start, KTO correction, and GRPO optimization into a scalable process that jointly enhances reasoning depth and computational efficiency.
    \item \textbf{High-quality datasets and strong empirical results.} We construct the EVA-SFT, EVA-KTO, and EVA-RL datasets to support stable training, achieving state-of-the-art performance across multiple video benchmarks.
\end{itemize}

\section{Related Works}
\label{sec:formatting}

\textbf{Agentic Video Understanding.}
Compared to traditional multimodal large language models (MLLMs) that treat the input video as static context~\citep{bai2025qwen25vltechnicalreport,feng2025videor1reinforcingvideoreasoning,maaz2024videochatgptdetailedvideounderstanding,zhang2025llavavideovideoinstructiontuning}, agentic video understanding methods enable MLLM-based agents to actively explore video content using external tools. According to the types of tools employed, existing approaches can be broadly divided into two categories. Ego-R1~\citep{tian2025ego} and M3-Agent~\citep{long2025seeing} leverage tools that assist in visual comprehension, such as invoking external MLLM APIs or conventional vision models, thereby depending heavily on the tool’s performance rather than the base model’s inherent multimodal capability.
The second category of works~\citep{xie2025video,fu2025love,meng2025openo3videogroundedvideo,he2025framethinker} equips MLLMs with sampling tools that extract partial or temporal visual information from the video. These methods primarily exploit the agent’s planning and recognition abilities, yet still treat the MLLM as a fixed component in a rigid workflow—receiving video input and generating predetermined parameters along a single dimension of control.
In contrast, our work restores true autonomy to the agent, enabling it not only to decide which parts of the video to observe, but also how to observe them, with flexible control over spatial resolution and temporal granularity.



\textbf{Tool-Integrated Reasoning Training.}
Equipping LLM-based agents with various external tools enables them to interact with the outside world~\citep{qin2023toolllmfacilitatinglargelanguage,li2025websailornavigatingsuperhumanreasoning,yan2024dependingshouldmentoringllm}, and even to autonomously generate and optimize complex workflows~\citep{zhang2025metaagentautomaticallyconstructingmultiagent,zhang2025aflowautomatingagenticworkflow}.
As foundation models have been trained to produce extended chains of thought for solving complex reasoning tasks~\citep{deepseekai2025deepseekr1incentivizingreasoningcapability,openai2024openaio1card}, recent studies~\citep{feng2025retoolreinforcementlearningstrategic,li2025searcho1agenticsearchenhancedlarge} have further integrated tool invocation into the reasoning process and optimized it through reinforcement learning.
In this work, we train an MLLM-based agent to iteratively plan and select informative frames, allowing it to flexibly adjust workflows according to the query and the visual content.
%
\section{Method}
\begin{figure*}
    
    \centering
    \includegraphics[width=0.8\linewidth]{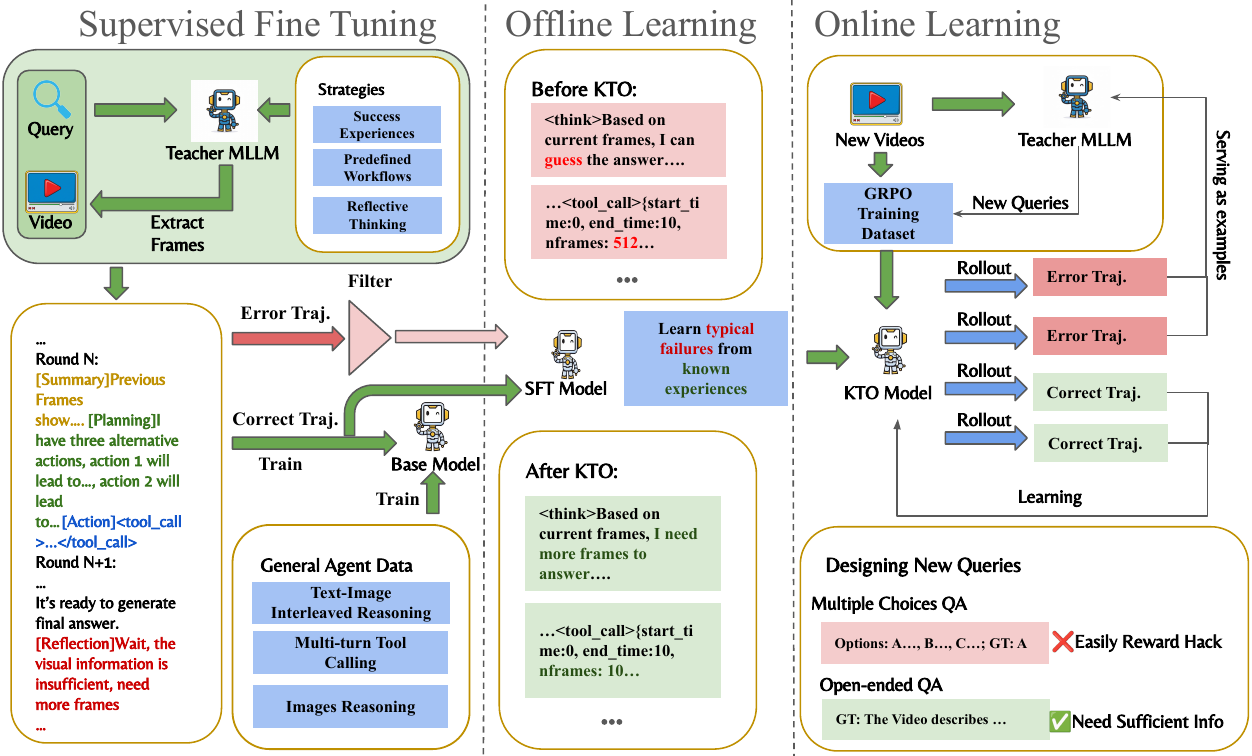}
    \caption{Data Pipeline and Training Stage of EVA. The base model is first fine-tuned on synthetic dataset with certain reasoning and tool-calling pattern. Then we use KTO to help the model learn from typical failures. Finally, we introduce a Data-Enhanced Multi-Stage GRPO training pipeline, where we collect the failure cases of current policy and employ an teacher MLLM to generate new open-ended video QA dataset.}
    \label{fig:fig2}
\end{figure*}
\subsection{Problem Setup}
We formulate the active video understanding problem as a Markov Decision Process (MDP). At each timestep \(t\), the agent observes a belief state:
\begin{equation}
\label{eq:grpo-states}
s_t = \{q,\,h_t,\,F_t\},
\end{equation}
where \(q\) denotes the user query, \(h_t\) represents the interleaved text–frame history, and \(F_t\) corresponds to the visual evidence (frames) obtained from tool calls. The policy of the agent is parameterized as \(\pi_\theta(a_t \mid s_t)\).

In video understanding tasks, answering a query does not always require observing uniformly sampled frames. In some cases, such frames are redundant, while in others they fail to provide sufficient evidence for correct reasoning—worse yet, presenting the full video upfront may mislead the planner by anchoring it to spurious or noisy visual cues~\citep{pan2025timesearch,fu2025love,wang2025video}. Therefore, at the initial step \(s_0\), the model is provided only with the query \(q\), without any visual information in our settings.  
To enable the agent to autonomously plan its use of visual tokens, we design a flexible \textit{frame-selection tool} that allows both temporal and spatial control.


\begin{center}
\begin{tabular}{ll}
\toprule
\textbf{Parameter} & \textbf{Description} \\
\midrule
$start\_time$ & Start of the temporal window \\
$end\_time$   & End of the temporal window \\
$nframes$     & Number of frames to sample \\
$resize$      & Spatial downsampling ratio \\
\bottomrule
\end{tabular}
\end{center}

The \textit{start\_time} and \textit{end\_time} specify the temporal window, while \textit{nframes} denotes the number of frames to sample within this interval. The \textit{resize} parameter enables flexible zoom-in and zoom-out operations.


Intuitively, selecting a larger number of frames enables the agent to better capture dynamic actions, while choosing a higher spatial resolution allows it to extract finer visual details from each frame. This tool schema provides a broad exploration space, encouraging the agent to learn how to allocate temporal and spatial information across rounds to derive precise answers.

Traditional agentic methods can be viewed as constrained instances of our proposed \textbf{EVA} framework. They typically employ fixed workflows—such as processing the entire video from the start—and offer limited action freedom (e.g., selecting only a time range). In contrast, EVA can not only execute these human-designed workflows but also dynamically adapt its plan based on the query and the extracted visual evidence, thereby enabling a more general and flexible paradigm for agentic video understanding.

Training such an end-to-end autonomous video agent requires not only visual comprehension and tool-use capabilities but also strong planning skills to determine \textit{what to watch} and \textit{when to answer} based on the question and available visual evidence. Hence, diverse high-quality datasets and efficient training strategies are crucial for developing such a general-purpose agent.


\subsection{Data Construction}

\paragraph{Supervised Fine-Tuning Data Pipeline}
We begin by employing Qwen2.5-VL-72B as the teacher MLLM and prompting it to generate high-quality agentic video understanding data following our problem setup. The source video QA pairs come from llava-video~\citep{zhang2025llavavideovideoinstructiontuning}, a short video QA dataset, and cgbench~\citep{chen2024cgbenchcluegroundedquestionanswering}, a long video QA dataset. To further enhance data diversity and quality, we design a variety of prompts to guide the teacher model. These include: \textit{Past Success Experiences} generated and summarized by the teacher MLLM itself; \textit{Diverse Workflow Hints} that instruct the model on how to plan and select frames efficiently; and \textit{Reflective Thinking Prompts} that encourage the model to carefully consider its actions.

Inspired by \citet{zhang2025agentlearningearlyexperience}, each SFT data instance follows the format: \textbf{Summary + Planning + Action + Reflection}. In the \textbf{Summary} stage, the MLLM generates a detailed description of the content for each frame, which explicitly pushes the model to attend to the returned visual evidence and thereby better ground its understanding of the tool’s parameters and outputs. During \textbf{Planning}, since an autonomous video agent possesses maximum flexibility to select actions from an extremely large action space, it is crucial to train its ability to propose potential actions based on current information while estimating their cost and outcome. In the \textbf{Action} stage, the model generates appropriate tool calls. Finally, as models tend to produce answers without sufficient visual evidence, resulting in degraded performance, we construct \textbf{Reflection} data that guide the model to evaluate whether the available visual information is adequate before producing an answer. If not, the model continues to call tools to gather additional information.


\paragraph{Kahneman-Tversky Optimization Data Pipeline}

The SFT-trained model effectively learns tool-calling formats and reasoning patterns; however, it still struggles to select appropriate strategies. Several typical failure cases share similar patterns: the model may generate answers without sufficient visual evidence, sample too many frames within a short temporal window, or too few frames across a relatively long one. To address these recurrent failures and stabilize subsequent online reinforcement learning, we employ the KTO framework.
Unlike DPO~\citep{rafailov2023direct}, which requires pairwise preference data and thus enforces a shared dialogue round—an assumption misaligned with our multi-turn interaction setting that may truncate strategies—KTO only requires single-sample preference labels (“chosen” or “rejected”). Compared with GRPO, KTO further enables the model to learn from externally collected experiences rather than self-play, leading to a more stable and sample-efficient training process.
Specifically, we collect incorrect trajectories from the SFT data construction pipeline and categorize them as rejected samples. The criterion of data selection are two folds. Firstly, we use LLM As Judge to select the trajectories whose reasoning process shows it do not have enough visual tokens but still generate a answer, representing for guessing. Secondly,  
We also resample high-quality successful trajectories as chosen data. This setup allows the model to learn fine-grained preferences between fully successful and failed trajectories. Figure~\ref{fig:fig2} illustrates representative examples before and after applying KTO.
\paragraph{GRPO Data Pipeline.}
GRPO is an online reinforcement learning framework in which the model generates multiple rollouts by itself and iteratively learns from both successes and failures. However, conventional GRPO typically relies on a static training dataset, and the model only iterates through it for a few epochs. This limitation becomes more pronounced when training a video agent. Traditional GRPO learns from failures based solely on a fixed query–video pair, which constrains the diversity of challenges encountered. For instance, the model may realize that its counting ability is weak, yet it can only improve from a limited set of failed queries and videos.

To address this issue, we introduce a Data-Enhanced GRPO pipeline. We first construct a reinforcement learning dataset by collecting failure cases from the KTO-trained model. After several GRPO training steps, we further gather new failure cases and use them as in-context examples for the teacher MLLM, which then generates new question–answer pairs for unseen videos from HD-VILA~\citep{xue2022advancing}, conditioned on those examples. And We will re-train GRPO model on the enhanced dataset.

Specifically, the teacher MLLM is prompted to produce open-ended QA pairs with concise answers. Compared to directly designing multiple-choice questions, this open-ended formulation mitigates reward hacking caused by answer guessing and offers a more efficient generation process for the teacher model, since designing balanced multiple-choice options often introduces unintended information cues and additional complexity.

\begin{figure}
    \centering
    \includegraphics[width=1\linewidth]{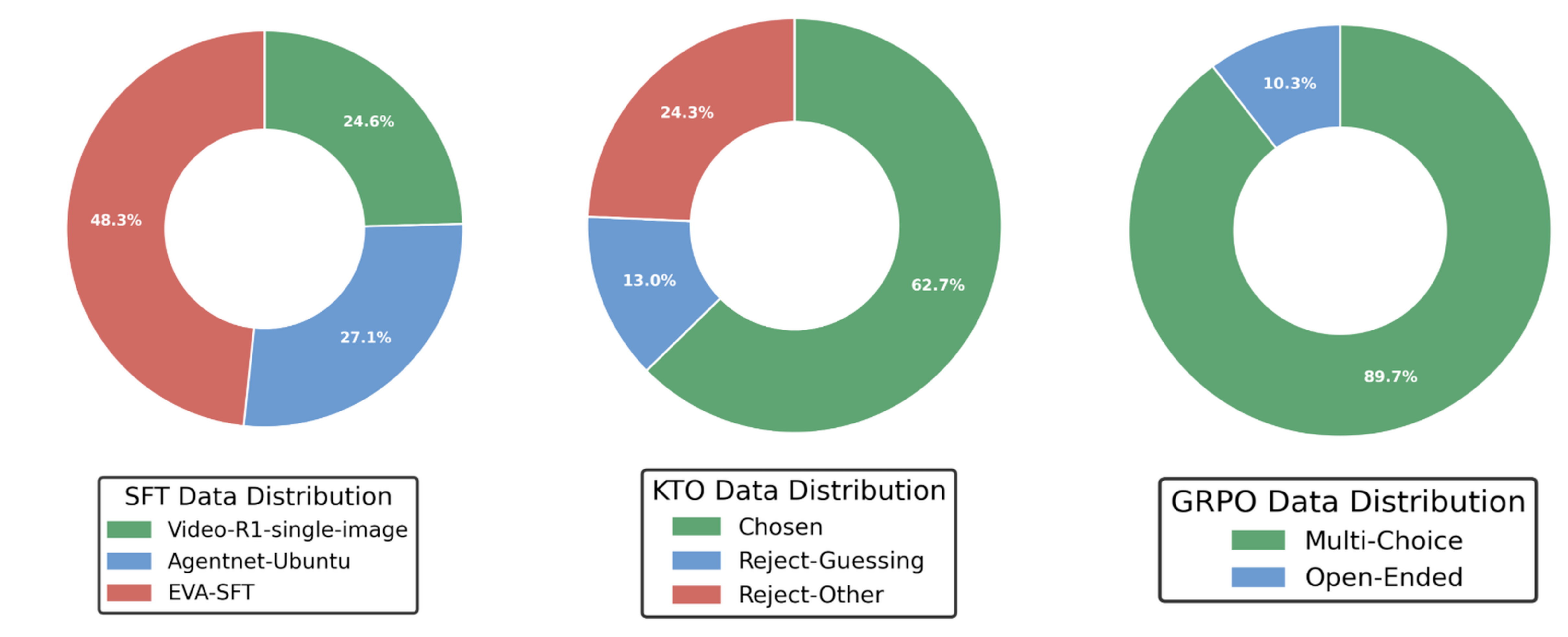}
    \caption{Distribution of the Training Dataset}
    \label{fig:3}
\end{figure}
\subsection{Reinforcement Learning}
We optimize the model via reinforcement learning using a mixed-format dataset. Specifically, we train EVA on both multiple-choice and open-ended questions. For multiple-choice questions, we adopt the Completeness Self-Verification (CSV) reward~\cite{pan2025timesearch} to ensure that EVA explicitly identifies the correct frames rather than relying on random guessing. For open-ended questions, we utilize the ROUGE score as the reward signal.

\subsubsection{GRPO Primer}
We employ \textit{Group Relative Policy Optimization} (GRPO)~\citep{shao2024deepseekmathpushinglimitsmathematical}, a KL-regularized policy optimization method that encourages high-return behaviors while constraining the policy to remain close to a reference model $\pi_{\mathrm{ref}}$ initialized via SFT and KTO. The training objective is formulated as:
\begin{equation}
\label{eq:grpo-obj}
\begin{aligned}
\max_{\theta}\;\; & \E_{\tau\sim\pi_\theta}\!\big[R(\tau)\big] 
 - \lambda\, \E_{(s,a)\sim\pi_\theta}\!
\left[\KL\!\left(\pi_\theta(\cdot\!\mid s)\,\|\,\pi_{\mathrm{ref}}(\cdot\!\mid s)\right)\right].
\end{aligned}
\end{equation}

\subsubsection{Reward Shaping}
\label{reward}

\paragraph{Accuracy Reward.}
Our GRPO training corpus contains both \textit{open-ended} and \textit{multiple-choice} question--answer pairs.  
Accordingly, we design a composite reward function that adapts to these two types of supervision:
\begin{equation}
\label{eq:reward-total}
\begin{aligned}
R(\tau)
= w_{\mathrm{acc}}\,r_{\mathrm{acc}}
 + w_{\mathrm{fmt}}\,r_{\mathrm{fmt}}.
\end{aligned}
\end{equation}

The accuracy reward is defined as:
\begin{equation}
\label{eq:reward-iscorrect}
r_{\mathrm{acc}} =
\begin{cases}
    r_{\mathrm{csv}}, & \text{if multiple-choice,} \\
    r_{\mathrm{rouge}}, & \text{if open-ended.}
\end{cases}
\end{equation}

Specifically, for multiple-choice tasks, we set up the same base model to act as a judge, feeding it the question alongside EVA's last round of retrieved images. We set $r_{\mathrm{csv}} = 1$ if and only if both the judge and EVA produce the correct answer; otherwise, $r_{\mathrm{csv}} = 0$. 

For open-ended tasks, let $R_1, R_2, R_L \in [0,1]$ denote the ROUGE-1, ROUGE-2, and ROUGE-L F1 scores between the generated answer and the ground truth (with stemming).  
The averaged ROUGE reward is then defined as:
\begin{equation}
\label{eq:reward-rouge}
r_{\mathrm{rouge}}
= \tfrac{1}{3}\big(R_1 + R_2 + R_L\big) \in [0,1].
\end{equation}

\paragraph{Format Reward.}
We also introduce a format reward to prevent the model from directly guessing the answer without proper reasoning. Specifically, if the model generates a tool call but ultimately yields an incorrect answer, we provide a compensatory format reward of $0.05$. Considering that the expected accuracy for random guessing is approximately $0.20$ or $0.25$ (depending on the number of choices), this deliberately low reward discourages the model from exploiting the formatting structure to gain undeserved points through random guessing.

\section{Experiments}

\subsection{Experiment Settings}

We choose Qwen2.5-VL-7B-Instruct as our base model, as it supports vision input at various resolution and save token when feeding frames with small resolutions. 

We first perform SFT using our EVA-SFT data and open-sourced agentic training data. The model is trained for 2 epoch, with batch size = 8 and learning rate = $2e{-6}$. The model is then trained using KTO~\citep{ethayarajh2024ktomodelalignmentprospect} using EVA-KTO data. Follow the recommended chosen and reject data proportion, there are 63\% correct trajectory and 37\% incorrect data in the training dataset. We keep the same learning rate and set $beta=0.1$.
We further train our model using GRPO based on our EVA-RL data, with 90\% open-ended QA and 10\% multiple-choice questions (MCQ). We use a combined reward as described in section \ref{reward}. In detail, multiple choice questions are rewarded for correct choice and open-ended questions are rewarded using rouge-score. The model is trained for 1 epoch, with batch size =64, number of rollout per sample =8 and learning rate $1e-6$, on 32 H100 GPUs.

We evaluate EVA on various video benchmarks including LSDBench~\citep{qu2025doesvisionlanguagemodellost}, LongVideoBench~\citep{wu2024longvideobenchbenchmarklongcontextinterleaved}, MLVU~\citep{zhou2025mlvubenchmarkingmultitasklong}, VideoMME~\citep{fu2025videommefirstevercomprehensiveevaluation}, LVBench~\citep{wang2025lvbenchextremelongvideo} and Video-Holmes~\citep{cheng2025videoholmesmllmthinklike}. 
The metrics for all benchmarks are accuracy, computed as the proportion of correctly predicted answers.
\subsection{Main Result}  
\paragraph{Sampling Dilemma Bench.}
 We first evaluate our model on the Sampling Dilemma Bench (LSDBench~\citep{qu2025doesvisionlanguagemodellost}) to examine its ability to balance sampling efficiency and visual understanding accuracy. As shown in Table~\ref{tab:tab1}, closed-source models such as Gemini-2.0-Flash achieve the highest accuracy (56.2\%) but rely on an extremely large number of visual tokens (over 700K), revealing the inefficiency of brute-force dense sampling. Among open-source models, Qwen2.5-VL and InternVideo2.5 achieve comparable results around 50\% using 256–768 frames. Building upon Qwen2.5-VL, we introduce an end-to-end video agent that performs planning-before-perception via iterative reasoning, tool calls, and reflection. This design allows the model to dynamically decide which frames to observe and reason over, rather than exhaustively processing all inputs. As shown in Table~\ref{tab:tab1}, EVA exhibits clear improvements, achieving 51.8\% with only 6.2K visual tokens, surpassing the baseline by +2.6\% while using significantly fewer tokens. These results demonstrate that our video agent effectively mitigates the sampling dilemma through reasoning-driven visual planning, enabling more efficient video understanding.
\paragraph{Long-Form Video Understanding.}
We further evaluate EVA on four long-form video benchmarks—LongVideoBench, MLVU, VideoMME, and LVBench—to assess its generalization in temporally extended scenarios. As shown in Table~\ref{tab:tab2}, EVA achieves strong and consistent results (55.1\%, 60.5\%, 59.9\% and 38.1\%), outperforming most open-source and adaptive agents while processing only about 20–30 frames per video. The number of frames is estimated by assuming 650 visual tokens per frame for fair comparison; in practice, EVA may use more frames with adaptively adjusted resolutions. This demonstrates the advantage of our planning-before-perception paradigm in long-form video understanding, where the video agent adaptively decides which segments to observe through iterative reasoning and reflection. By dynamically allocating attention rather than relying on fixed dense sampling, EVA maintains high accuracy with minimal visual tokens, effectively addressing the long-context challenge in video understanding.

\paragraph{Zero-Shot Performance on Video Reasoning Bench.}
We further evaluate EVA on the Video-Holmes benchmark, which assesses diverse reasoning abilities. As shown in Table~\ref{tab:tab3}, despite being evaluated in a zero-shot setting, EVA achieves competitive overall performance (32.6\%, 32.9\% and 36.7\% for EVAs), comparable to open-source models such as Video-R1 and VideoChat-R1 with uniformly sampled frame schema. This demonstrates the strong transferability of our reasoning-driven video agent: even without task-specific supervision, EVA can generalize to multi-step reasoning and causal understanding across long temporal contexts. The results highlight that our planning-before-perception paradigm not only improves efficiency but also enables robust general reasoning in video understanding.

\begin{table}[]
\caption{Performance on sampling dilemma bench. We report different stages of EVA model's performance on LSDbench. SOTA model performance are directly from~\citep{he2025framethinker}. The visual token are roughly measured by 258, 144 ,256 and 650 token per frame for gemini, LongVA, LongVila and Qwen-VL family models.}
\raggedright
\resizebox{0.46\textwidth}{!}{
\begin{tabular}{lccc}
\hline
\textbf{Method}                         & \multicolumn{1}{l}{\textbf{Frames}} & \multicolumn{1}{l}{\textbf{Visual Token}} & \multicolumn{1}{l}{\textbf{Acc (\%)}} \\
\hline
\multicolumn{4}{l}{\cellcolor[HTML]{EBECED}\textbf{Close Source Model}}                                                                                           \\
{\color[HTML]{8F959E} Gemini-2.0-Flash~\cite{team2024gemini}} & {\color[HTML]{8F959E} 2700}         & {\color[HTML]{8F959E} 696.6k}             & {\color[HTML]{8F959E} 56.2}           \\
\hline
\multicolumn{4}{l}{\cellcolor[HTML]{EBECED}\textbf{Open Source Model}}                                                                                            \\
                                        & 256                                 & 36.9k                                        & 31.3                                  \\
                                        & 512                                 & 73.7k                                         & 33.0                                  \\
\multirow{-3}{*}{LongVA~\citep{zhang2024long}}                & 1024                                & 147.5k                                         & 32.5                                  \\
Qwen2-VL~\citep{wang2024qwen2}                                & 256                                 & 166.4k                                         & 48.0                                  \\
                                        & 256                                 & 166.4k                                         & 50.1                                  \\
\multirow{-2}{*}{Qwen2.5-VL~\citep{bai2025qwen2}}            & 768                                 & 499.2k                                         & 52.5                                  \\
LongVila~\citep{chen2024longvila}                                & 256                                 & 65.5k                                        & 49.8                                  \\
Qwen2.5-VL(RHS)~\citep{he2025framethinker}                         & 225                                 & 146.2k                                         & 52.2                                  \\
\hline
\multicolumn{4}{l}{\cellcolor[HTML]{E9FBF9}\textbf{Baseline}}                                                                                                     \\

                                        & 16                                  & 10.5k                                     & 47.7                                  \\
\multirow{-2}{*}{Qwen2.5-VL*}           & 32                                  & 21.0k                                     & 49.2                                  \\
\hline
\multicolumn{4}{l}{\cellcolor[HTML]{E9FBF9}\textbf{Ours}}                                                                                                         \\

EVA                              & 76.9                                & 10.3k                                     & \textbf{51.0}                  \\
\hline
\end{tabular}}

\label{tab:tab1}
\end{table}

\begin{table*}[t] 
\caption{Main performance on multiple video understanding benchmark. Baseline results are directly cited from~\citep{he2025framethinker}.The number of frames for EVA, which is indicated by *, is estimated by assuming 650 visual tokens per frame for fair comparison; the actual number of frames may vary depending on the resolution determined by the model adaptively.} 
\centering
\begin{tabular}{lcccccccc}
\hline
                                      & \multicolumn{2}{c}{\textbf{LongVideoBench}}             & \multicolumn{2}{c}{\textbf{MLVU}}                            & \multicolumn{2}{c}{\textbf{VideoMME-Long/Overall}}               & \multicolumn{2}{c}{\textbf{LVBench}}                      \\
\multirow{-2}{*}{\textbf{Model}}      & \textbf{Frame}            & \textbf{Acc}                & \textbf{Frame}                 & \textbf{Acc}                & \textbf{Frame}                & \textbf{Acc}                     & \textbf{Frame}              & \textbf{Acc}                \\
\hline
\multicolumn{9}{l}{\cellcolor[HTML]{F8F9FA}\textbf{Close Source Models}}                                                                                                                                                                                                                      \\
{\color[HTML]{8F959E} GPT-4o~\cite{hurst2024gpt}}         & {\color[HTML]{8F959E} 32} & {\color[HTML]{8F959E} 58.2} & {\color[HTML]{8F959E} 0.5 fps} & {\color[HTML]{8F959E} 64.6} & {\color[HTML]{8F959E} 384}    & {\color[HTML]{8F959E} 65.3/71.9} & {\color[HTML]{8F959E} 60}   & {\color[HTML]{8F959E} 48.9} \\
{\color[HTML]{8F959E} Gemini-1.5-Pro~\cite{team2024gemini}} & {\color[HTML]{8F959E} 32} & {\color[HTML]{8F959E} 55.2} & {\color[HTML]{8F959E} -}       & {\color[HTML]{8F959E} -}    & {\color[HTML]{8F959E} 0.5fps} & {\color[HTML]{8F959E} 67.4/75.0} & {\color[HTML]{8F959E} 3600} & {\color[HTML]{8F959E} 33.1} \\
\hline
\multicolumn{9}{l}{\cellcolor[HTML]{F8F9FA}\textbf{Static Frame Sampling}}                                                                                                                                                                                                                    \\
ShareGPT4Video~\citep{chen2024sharegpt4video}                        & 16                        & 39.7                        & 16                             & 46.4                        & 16                            & 35.0/39.9                        & -                           & -                           \\
LongVA~\citep{zhang2024long}                                & -                         & -                           & 256                            & 56.3                        & 128                           & 46.2/52.6                        & -                           & -                           \\
VITA-1.5-7B~\citep{fu2025vita}                           & -                         & -                           & -                              & -                           & 16                            & 47.1/56.1                        & -                           & -                           \\
Video-R1~\citep{feng2025videor1reinforcingvideoreasoning}                              & 32                        & 52.7                        & 32                             & 60.2                        & 32                            & 49.4/59.9                        & 32                          & 35.3                        \\
VideoChat-R1~\citep{li2025videochat}                          & 32                        & 49.1                        & 32                             & 54.3                        & 32                            & 46.2/-                           & 32                          & 34.3                        \\
Qwen2.5-VL~\citep{bai2025qwen2}                            & 32                        & 43.2                        & 32                             & 48.4                        & 32                            & 44.7/53.6                        & 32                          & 31.6                        \\
\hline
\multicolumn{9}{l}{\cellcolor[HTML]{F8F9FA}\textbf{Adaptive Agent}}                                                                                                                                                                                                                           \\
VideoAgent~\citep{fan2024videoagent}                            & -                         & -                           & -                              & -                           & 87                            & 49.0/56.0                        & 25.5                        & 29.3                        \\
FrameThinker~\citep{he2025framethinker}                          & 21.1                      & 52.9                        & 23.2                           & 59.1                        & 24.1                          & 47.6/-                           & 23.9                        & 36.6                        \\
VideoMTR~\citep{xie2025video}                              & -                         & -                           & -                              & -                           & 32                            & 51.0/59.0                        & -                           & -                           \\
\hline
\multicolumn{9}{l}{\cellcolor[HTML]{E9FBF9}\textbf{Ours}}                                                                                                                                                                                                                                     \\
EVA-SFT                                   & 33.8*                     & 49.9                      &  46.7*                              &           52.3                  & 26.6*                         & 45.8/56.0                        &   56.2*                          & 26.5     \\
EVA-KTO                                   & 35.6*                     & 53.2                        &  28.7*                              &           57.4                & 24.1*                         & 45.1/56.5                       &   34.5*                          & 36.0     \\
EVA-GRPO                                   & 25.3*                     & \textbf{55.0}                        &  22.2*                              &           \textbf{68.3}                  & 22.8*                         & \textbf{48.4/60.2}                        &   26.8*                          & \textbf{43.3}     \\
\hline
\end{tabular}
\label{tab:tab2}
\end{table*}

\begin{table*}[t] 
\centering
\caption{Zero-shot performance on video reasoning benchmark: Video-Holmes~\citep{cheng2025videoholmesmllmthinklike}, where SR stands for Social Reasoning; IMC stands for Intention \& Motive Chaining; TCI stands for Temporal Causal Inference; TA Timeline Analysis; MHR stands for Multimodal Hint Reasoning; PAR stands for Physical Anomaly Reasoning; CTI stands for Core Theme Inference.}
\begin{tabular}{lccccccccc}
\hline
Model                                   & \multicolumn{1}{l}{Frame} & \multicolumn{1}{l}{SR} & \multicolumn{1}{l}{IMC} & \multicolumn{1}{l}{TCI} & \multicolumn{1}{l}{TA} & \multicolumn{1}{l}{MHR} & \multicolumn{1}{l}{PAR} & \multicolumn{1}{l}{CTI} & \multicolumn{1}{l}{Overall} \\
\hline
\multicolumn{10}{l}{\cellcolor[HTML]{F8F9FA}\textbf{Close Source Models}}                                                                                                                                                                                                             \\
{\color[HTML]{8F959E} GPT-4o~\cite{hurst2024gpt}}           & 32                        & 50.0                   & 49.6                    & 38.8                    & 30.0                   & 44.0                    & 39.2                    & 37.0                    & 42.0                        \\
{\color[HTML]{8F959E} Gemini-2.0-Flash~\cite{team2024gemini}} & -                         & 41.8                   & 33.7                    & 23.1                    & 20.5                   & 30.1                    & 26.8                    & 33.7                    & 30.6                        \\
\hline
\multicolumn{10}{l}{\cellcolor[HTML]{F8F9FA}\textbf{Open Source Model}}                                                                                                                                                                                                               \\
InternVL2.5-8B~\citep{chen2024expanding}                          & 32                        & 28.0                   & 32.2                    & 21.5                    & 7.7                    & 25.7                    & 23.8                    & 22.6                    & 23.8                        \\
InternVL3-8B~\citep{zhu2025internvl3}                            & 32                        & 29.5                   & 40.7                    & 37.9                    & 35.1                   & 24.6                    & 38.9                    & 24.1                    & 32.3                        \\
Qwen2.5-VL-7B~\citep{bai2025qwen2}                           & 32                        & 38.4                   & 34.8                    & 17.6                    & 30.0                   & 27.1                    & 18.6                    & 25.2                    & 27.8                        \\
SEED-Bench-R1~\citep{chen2025exploring}                           & 32                        & 42.8                   & 35.1                    & 25.6                    & 40.5                   & 29.2                    & 29.9                    & 32.6                    & 33.5                        \\
VideoChat-R1~\citep{li2025videochat}                            & 32                        & 42.1                   & 38.8                    & 24.5                    & 39.5                   & 29.5                    & 27.8                    & 29.3                    & 33.0                        \\
Video-R1~\citep{feng2025videor1reinforcingvideoreasoning}                                & 32                        & 48.6                   & \textbf{41.7}                    & \textbf{28.9}                    & 34.5                   & \textbf{31.0}                    & 33.6                    & 35.6                    & 36.5                        \\
\hline
\multicolumn{10}{l}{\cellcolor[HTML]{E9FBF9}\textbf{Ours}} \\
EVA-SFT  & 11.5* & 44.5 & 33.7 & 26.4 & 39.5 & 23.2 & 31.9 & 32.2 & 32.6 \\
EVA-KTO  & 5.8*  & 48.6 & 36.2 & 22.7 & 39.5 & 22.9 & 32.0 & 31.1 & 32.9 \\
EVA-GRPO & 36.8* & \textbf{49.3} & 39.5 & \textbf{30.4} & \textbf{44.5} & 27.1 &   37.6 & \textbf{35.2} & \textbf{37.2} \\

\hline
\end{tabular}
\label{tab:tab3}
\end{table*}

\subsection{Ablation Study}
\begin{figure}
    \centering
    \includegraphics[width=1.0\linewidth]{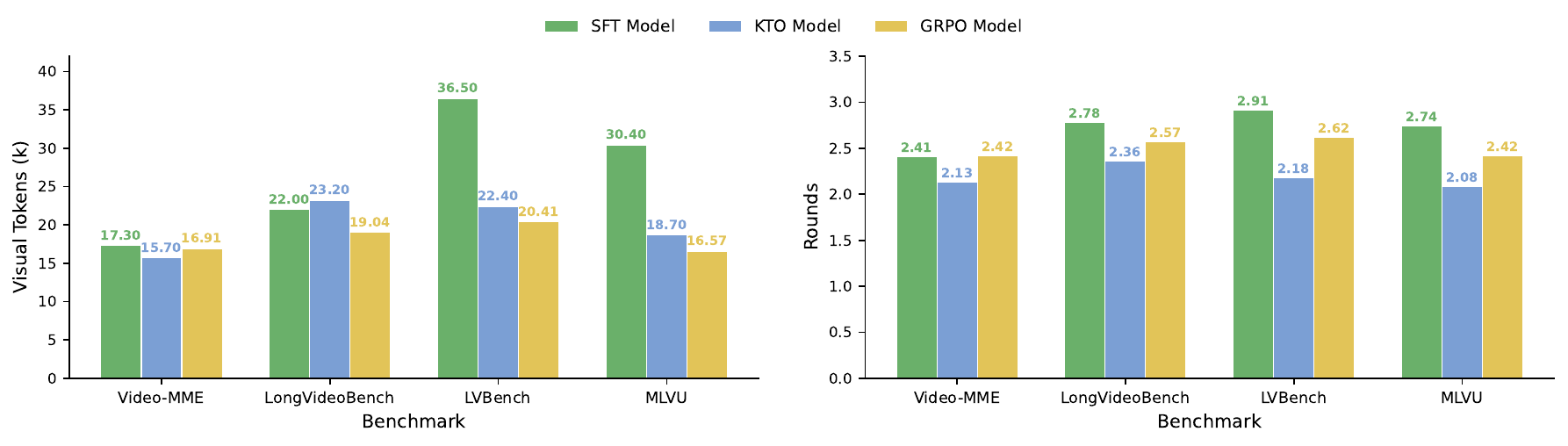}
    \caption{Distribution of Rounds and Visual Token cross Models and Benchmarks}
    \label{fig:round_dis}
\end{figure}
\begin{figure}[t]
    \centering
    \begin{subfigure}{0.48\linewidth}
        \centering
        \includegraphics[width=\linewidth]{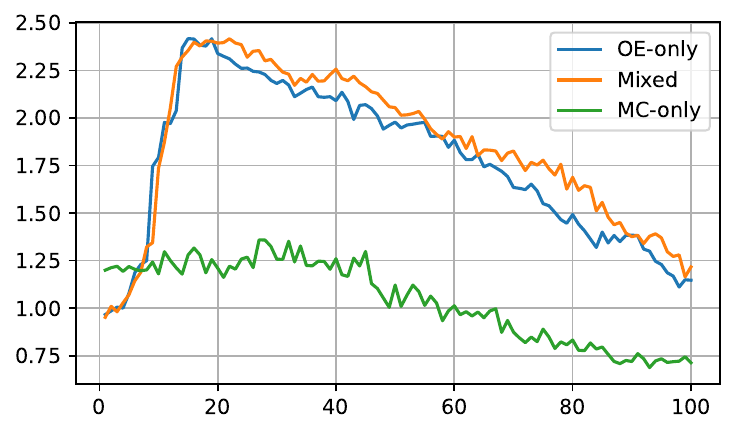}
        \caption{}
        \label{fig:entropy-a}
    \end{subfigure}
    \hfill
    \begin{subfigure}{0.48\linewidth}
        \centering
        \includegraphics[width=\linewidth]{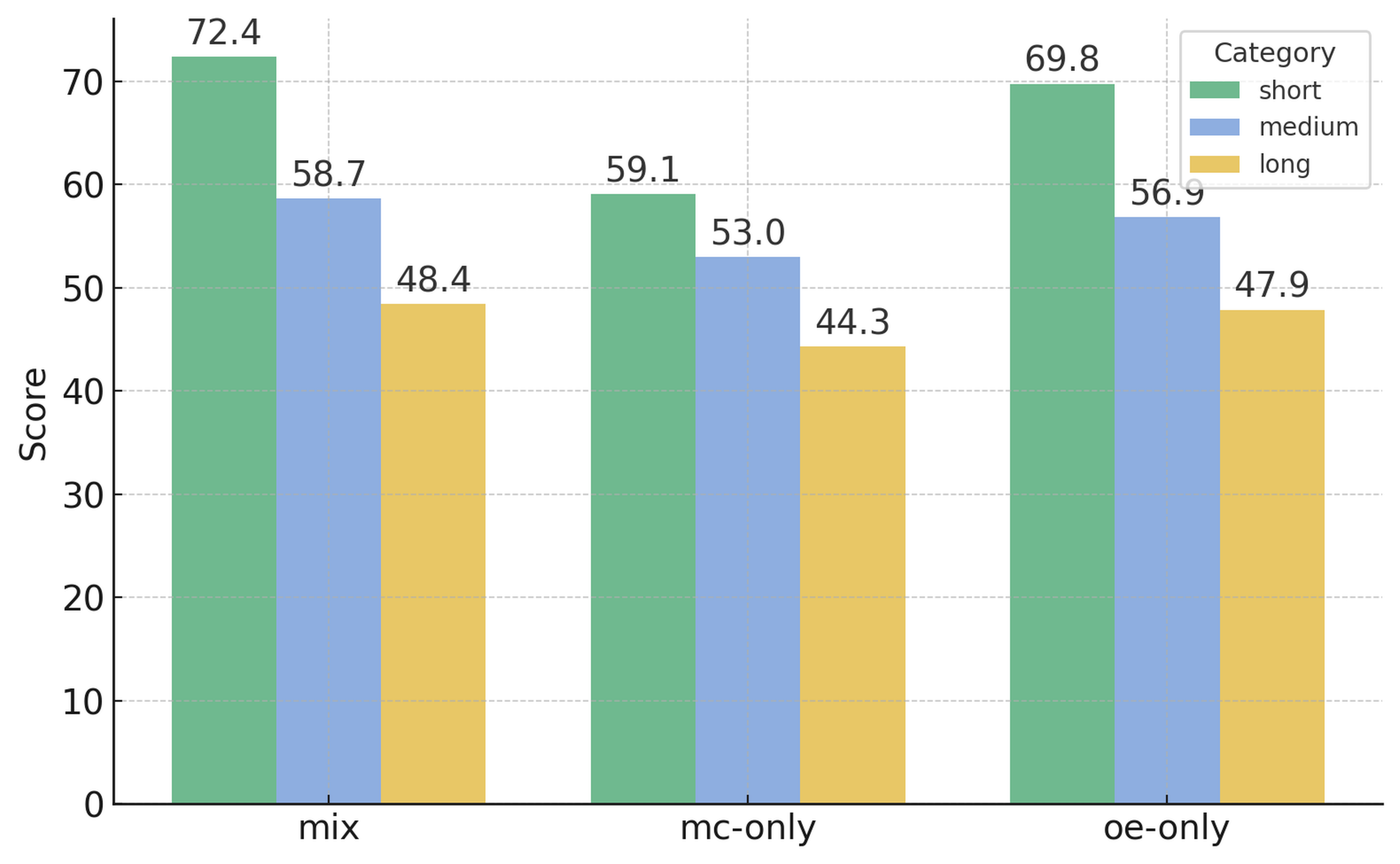}
        \caption{}
        \label{fig:entropy-b}
    \end{subfigure}

    \caption{Ablation study on the GRPO training dataset. The comparison between multi-choice (MC) only, open-ended (OE) only, and mixed (MC+OE) data shows that mixed data provides a more effective learning environment for the agent, which leads to better performance on VideoMME.}
    \label{fig:entropy}
\end{figure}

\paragraph{Training Schema}
Our ablation study confirms the SFT–KTO–GRPO sequence provides a clear evolutionary path for EVA. As shown in Fig.~\ref{fig:round_dis} and Table~\ref{tab:tab2}, the SFT model consumes a large number of frames and rounds yet achieves the lowest performance, indicating that supervised fine-tuning alone teaches the agent to follow tool-calling formats but not to explore videos efficiently. The KTO model significantly reduces both frame consumption and interaction rounds while delivering a substantial improvement over SFT. Interestingly, the GRPO model further reduces the number of sampled frames compared to KTO, yet increases the number of interaction rounds, and achieves the highest scores across all benchmarks. This reveals a shift in exploration strategy: rather than passively consuming fewer frames in fewer steps, the GRPO-trained agent learns to engage in more deliberate, multi-round reasoning while allocating its visual token budget more precisely in each round. These results illustrate that our training scheme progressively transforms the Video Agent from a format-following imitator into a strategic video explorer that actively plans when and where to look.
\paragraph{Data Composition in GRPO}
Although KTO effectively encourages broader exploration, we observe that the model may still hack the reward by producing plausible but visually unsupported guesses when limited visual evidence is available. To mitigate this issue, we introduce open-ended data during GRPO, forcing the agent to ground its answers in the visual content even when exploring diverse trajectories. As illustrated in Fig.~\ref{fig:entropy}, mixing open-ended data with task-specific data leads to more stable training and clearly improved performance, demonstrating that proper data composition is crucial for preventing reward hacking and ensuring visually consistent reasoning.

\subsection{Computation Efficiency Compared with Uniform Sampling Schema}
Although our agent performs multi-round planning and perception, its overall computation remains highly efficient. As shown in Table~\ref{tab:tab1}, the total number of tokens is comparable to or even lower than static uniform-sampling baselines, with text tokens occupying only a small fraction of the computation budget. Consequently, the inference runtime does not scale with the number of reasoning steps but is instead dominated by the compact set of adaptively selected visual tokens. This allows our agent to achieve competitive or superior efficiency compared with uniform sampling, despite operating under a more expressive reasoning framework.

\subsection{Case Study}
\begin{figure}
    \centering
    \includegraphics[width=1\linewidth]{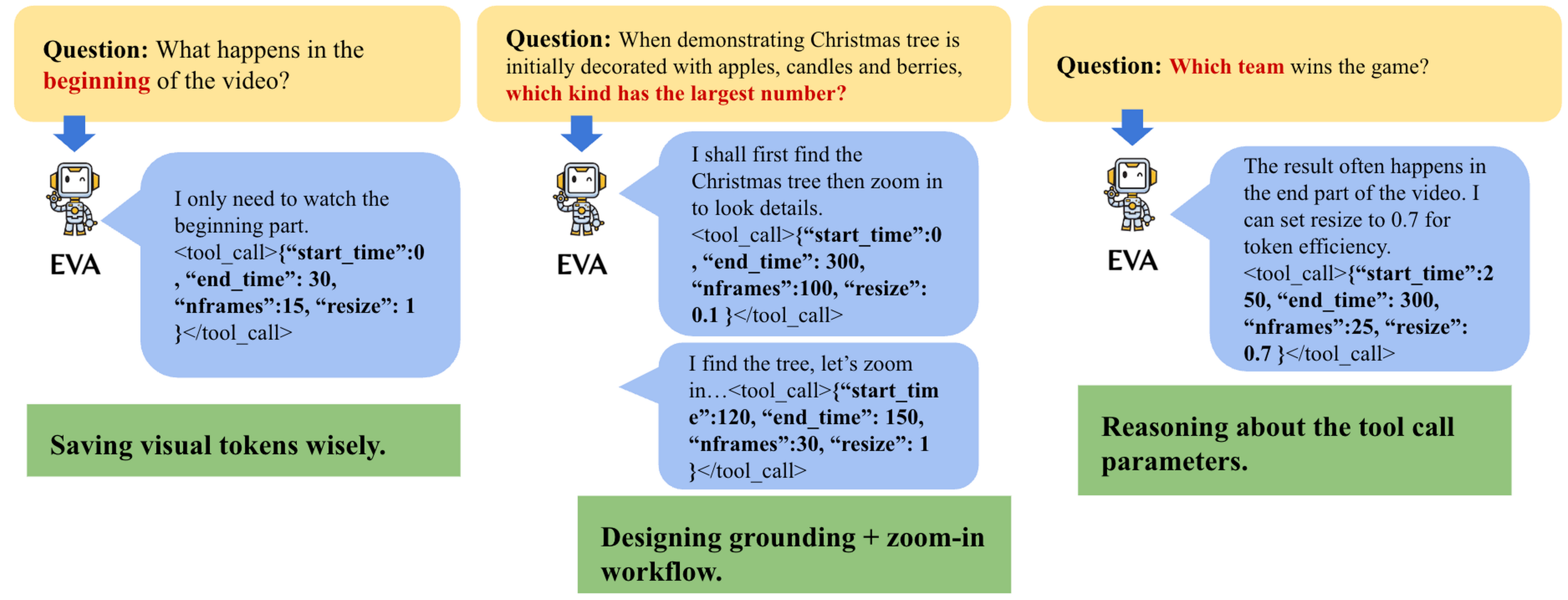}
    \caption{Diverse Workflows generated by EVA.}
    \label{fig:case}
\end{figure}

Figure~\ref{fig:case} illustrates the diverse workflows produced by EVA for different types of questions. Many queries require visual information from only a specific segment of the video (e.g., the beginning or the end). Under the plan-before-perception scheme, EVA allocates visual tokens precisely where needed, avoiding redundant usage that often occurs in traditional MLLM or agent methods, which perceive before reasoning or tool-calling. 

When a query genuinely requires extensive visual evidence, EVA can also behave like conventional video agents—first sampling uniformly for grounding and then zooming in on key frames to extract finer details, showing EVA is a more general video agent.

The larger action space further differentiates EVA from existing Video Agents. For example, when performing a zoom-in operation, prior agents can only adjust the temporal range but cannot vary both the number of sampled frames and the spatial resolution. In contrast, EVA autonomously selects all relevant parameters, enabling workflows that not only cover those of other agents but also surpass them through more efficient visual token usage.

\section{Conclusion and Limitation
}
In this work, we take a step toward building truly autonomous video-understanding agents by introducing a query-driven framework that integrates understanding, planning, tool use, and reflection in an iterative loop. Through a three-stage training paradigm—SFT cold-start, KTO, and GRPO with a commonly used reward—the model learns to balance perceptual efficiency and reasoning depth, progressively evolving from a passive video recognizer into an adaptive and self-directed agentic watcher.
Despite these promising results, our approach still faces several limitations. The current reasoning loop relies on pre-defined tool interfaces and may struggle with unseen or noisy query distributions. Future work will explore more flexible tool ecosystems, self-evolving reasoning strategies, and cross-modal memory mechanisms to further enhance autonomy and generalization in long-horizon video understanding.


\clearpage

\newpage
{
    \small
    \bibliographystyle{ieeenat_fullname}
    \bibliography{main}
}
\section{Appendix}
\begin{figure*}[t]
    \centering
    \includegraphics[width=1\linewidth]{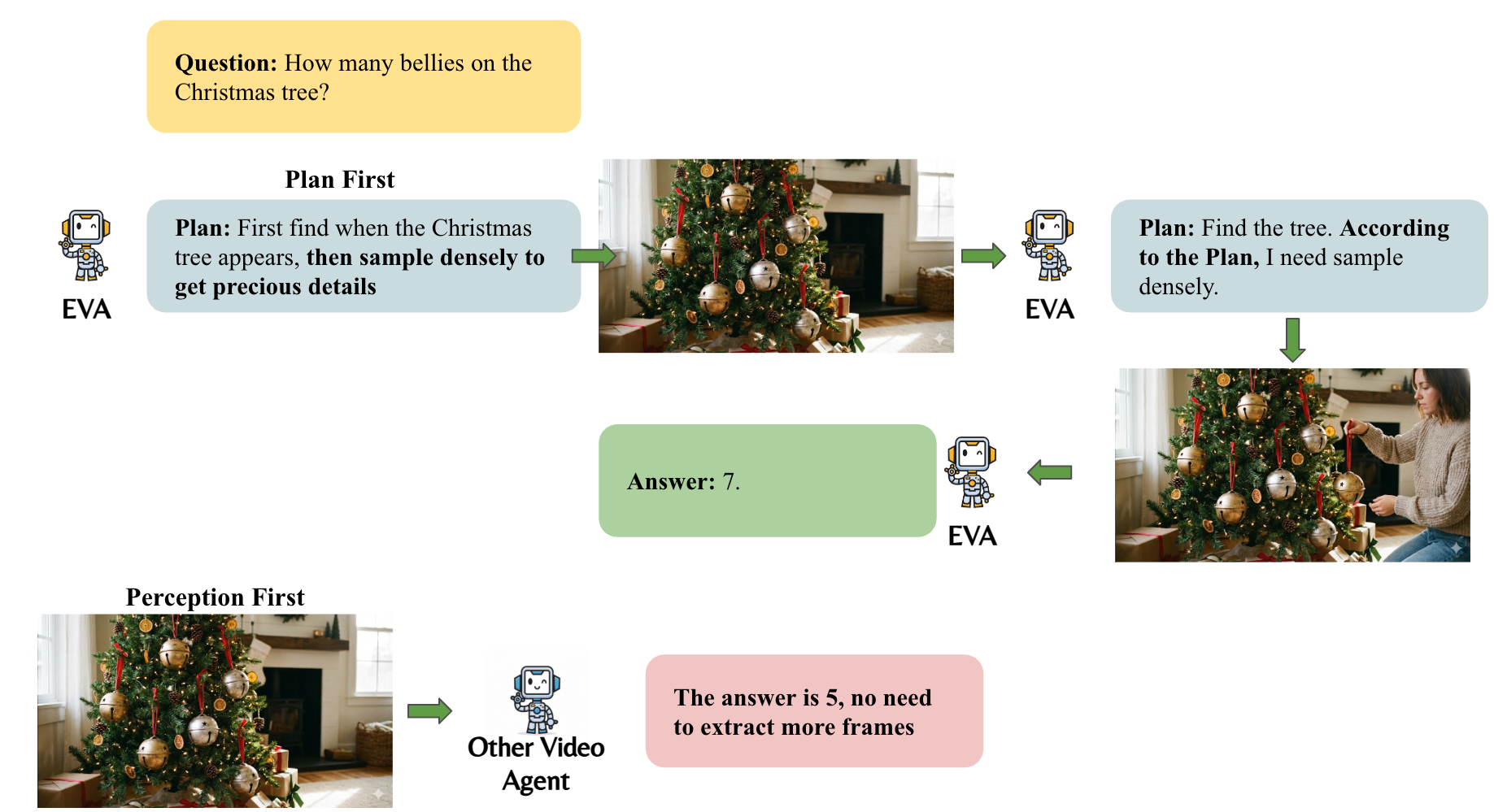}
    \caption{The advantage of plan before perception matters}
    \label{fig:perceptionfirst}
\end{figure*}
\subsection{Why plan before perception}
The core philosophy that distinguishes EVA from prior video-agent systems is its planning-before-perception paradigm. This new agentic video-understanding framework offers several key advantages.

\paragraph{Avoiding visual misguidance.}
Under the traditional perception-first paradigm, uniformly sampled frames often contain irrelevant or noisy actions that mislead the model during reasoning. In contrast, planning-first allows the agent to construct a textual plan that clarifies its intent before interacting with the video. This plan serves as a guiding prior, steering the agent toward the visual evidence that is truly required by the question and preventing distraction from irrelevant content.
As illustrated in Figure~\ref{fig:perceptionfirst}, EVA first formulates a plan that guides its subsequent observation process, enabling it to inspect the video with explicit purpose. Other video agents, however, are forced to rely solely on uniformly sampled frames, which frequently introduce sampling bias. Such bias can distort the agent’s perception and lead to incorrect conclusions. Without an explicit plan, even agents equipped with frame-selection tools cannot reliably decide when and how to use them.

\paragraph{Saving visual tokens.}
For long videos, sampling the entire content at high resolution is prohibitively expensive. In many cases, only a small temporal segment or a low-resolution preview is sufficient for answering the query. Planning-before-perception naturally reduces visual-token usage by enabling the agent to identify which parts of the video matter, thereby improving both efficiency and accuracy.

\paragraph{Active perception rather than passive observation.}
Traditional perception–reasoning pipelines inherently operate under a passive observation regime, where the model is restricted to whatever frames are provided to it and therefore lacks the ability to control what visual evidence should be acquired. Such passivity fundamentally limits reasoning: when the observation is fixed, the model’s understanding is constrained by noise, sampling bias, and irrelevant visual content. In contrast, agentic intelligence requires an active perceptual process in which the system explicitly determines what information is necessary, decides how to obtain it, and selectively interacts with the environment through tool calls to gather targeted visual evidence. The planning-before-perception paradigm enables precisely this mode of active perception. By formulating an explicit plan prior to observing the video, the agent first establishes a hypothesis about what information is needed for solving the task, and then acquires only the relevant content to verify or refine this hypothesis. This deliberative loop—intention formation followed by targeted perception—allows the agent to transcend passive frame consumption and instead engage in purposeful, goal-driven visual information gathering, which is essential for robust and scalable video understanding.
\subsection{Data Pipeline Details}

\begin{figure*}
        \centering
        \includegraphics[width=1\linewidth]{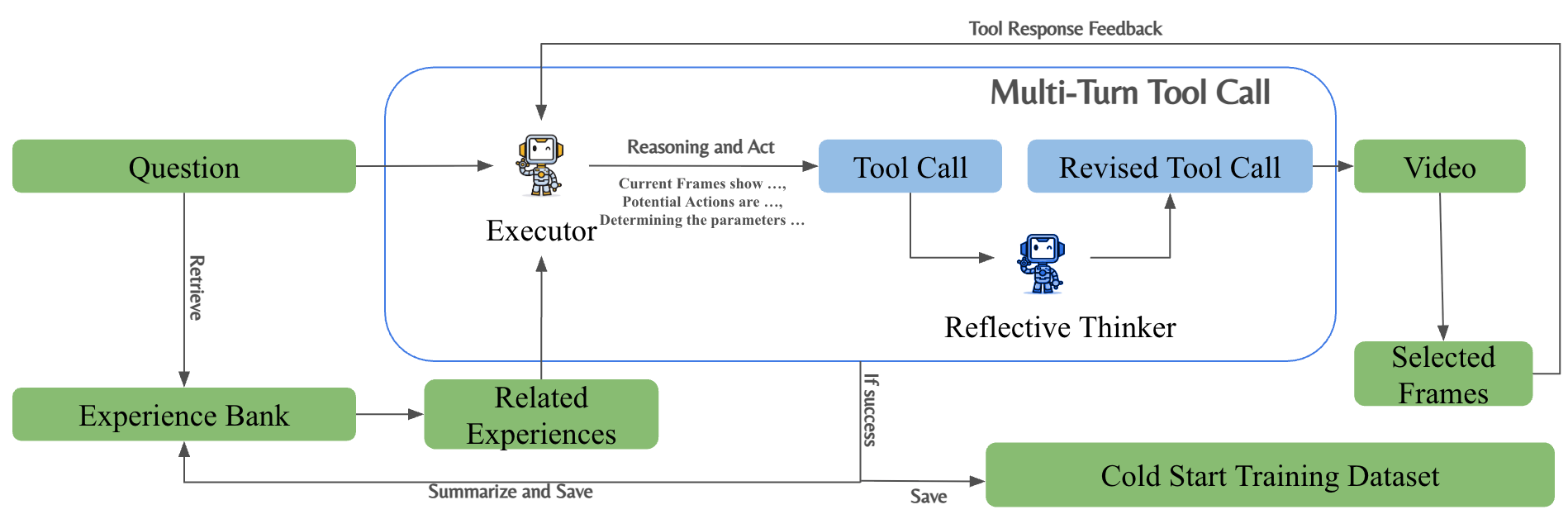}
        \caption{Cold Start Data Pipeline}
        \label{fig:sftpipeline}
\end{figure*}
We developed a Multi-Agent Data Pipeline to generate a high-quality supervised fine-tuning dataset, as detailed in Figure \ref{fig:sftpipeline}. Upon receiving a query, an Executor agent analyzes the current context—consisting of the initial query and, in subsequent rounds, relevant video frames—to evaluate potential actions and their expected outcomes. The Executor then optimizes action parameters and determines whether sufficient information exists to submit a final answer. Proposed tool calls are further scrutinized by a Reflective Thinker, which assesses the reasonableness of the parameters. Following this multi-turn loop, successful trajectories are archived in an Experience Bank. These trajectories are retrieved based on query similarity to guide the Executor, thereby enhancing its success rate in future iterations.

\subsection{EVA Behavior Analysis}
\begin{figure*}[t]
    \centering
    \begin{subfigure}[b]{0.48\linewidth}
        \centering
        \includegraphics[width=\linewidth]{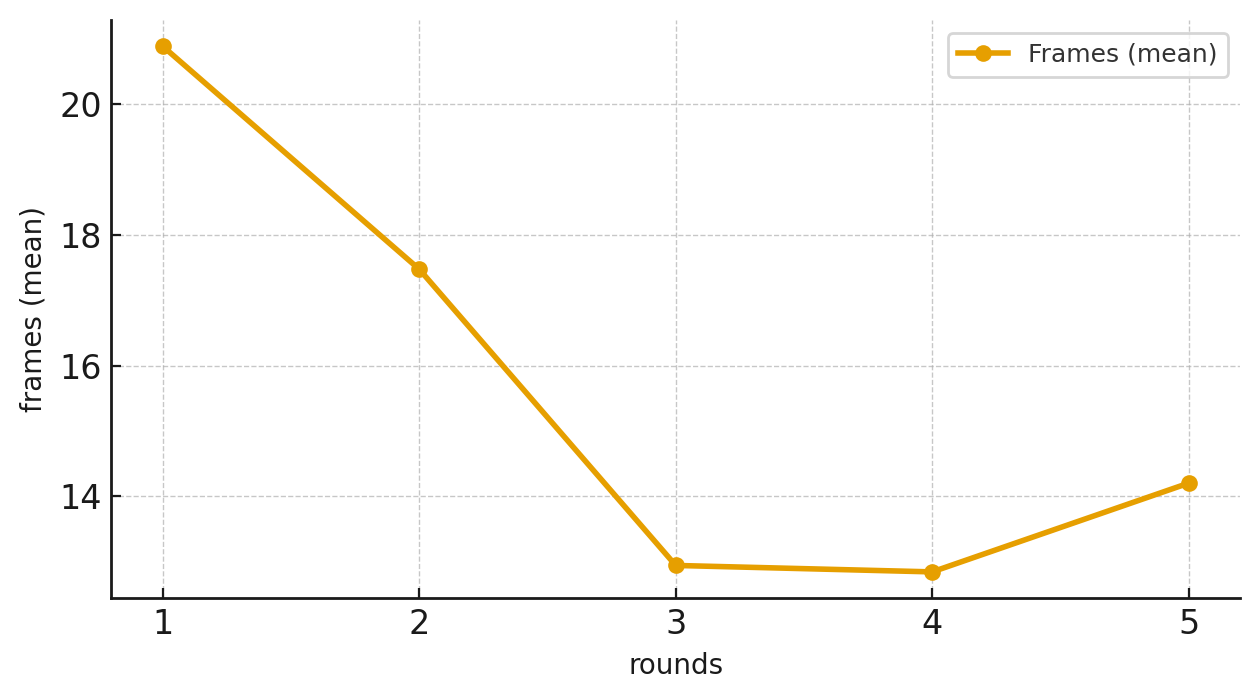}
        \caption{Average Frames (nframe*fps)}
        \label{fig:eva_frames}
    \end{subfigure}
    \hfill
    \begin{subfigure}[b]{0.48\linewidth}
        \centering
        \includegraphics[width=\linewidth]{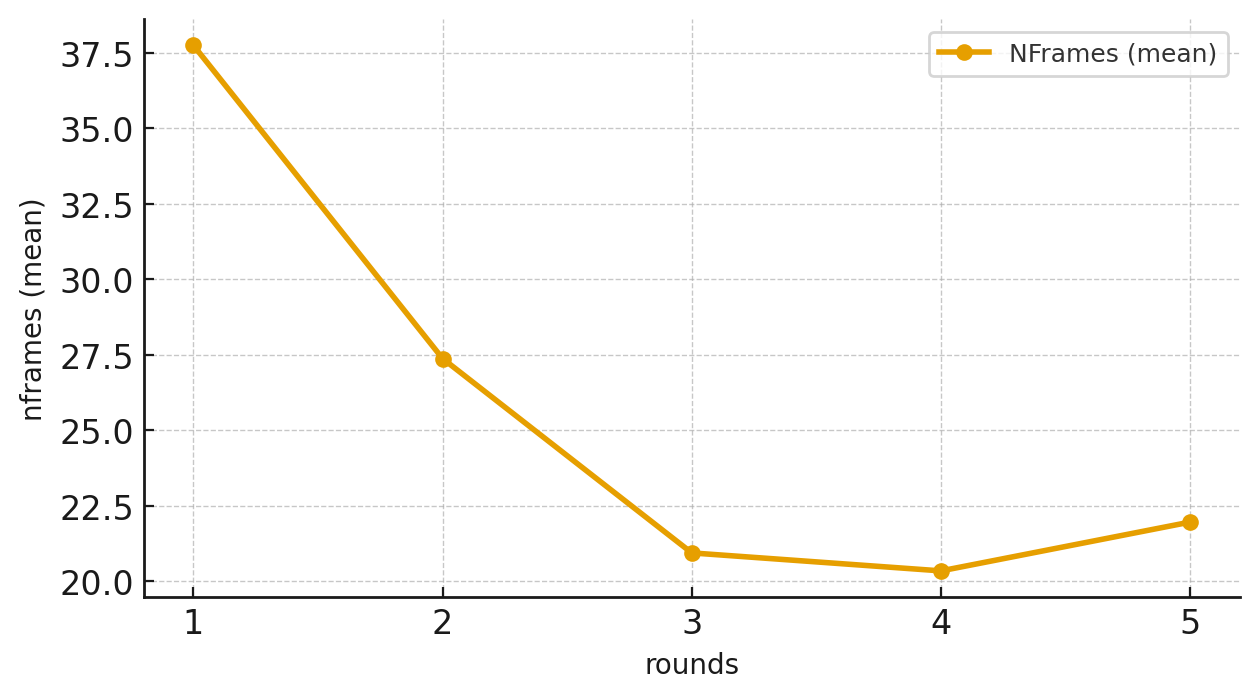}
        \caption{Average nframe}
        \label{fig:eva_nframe}
    \end{subfigure}

    \vspace{0.3cm}

    \begin{subfigure}[b]{0.48\linewidth}
        \centering
        \includegraphics[width=\linewidth]{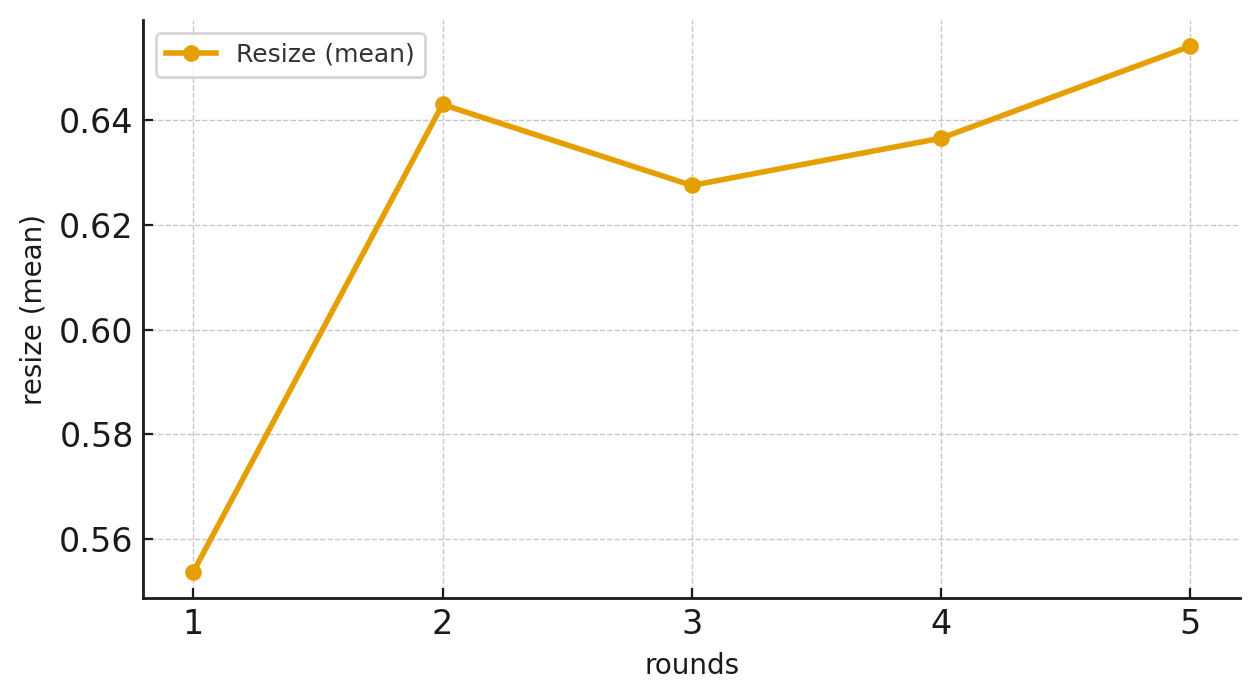}
        \caption{Average resize}
        \label{fig:eva_resize}
    \end{subfigure}
    \hfill
    \begin{subfigure}[b]{0.48\linewidth}
        \centering
        \includegraphics[width=\linewidth]{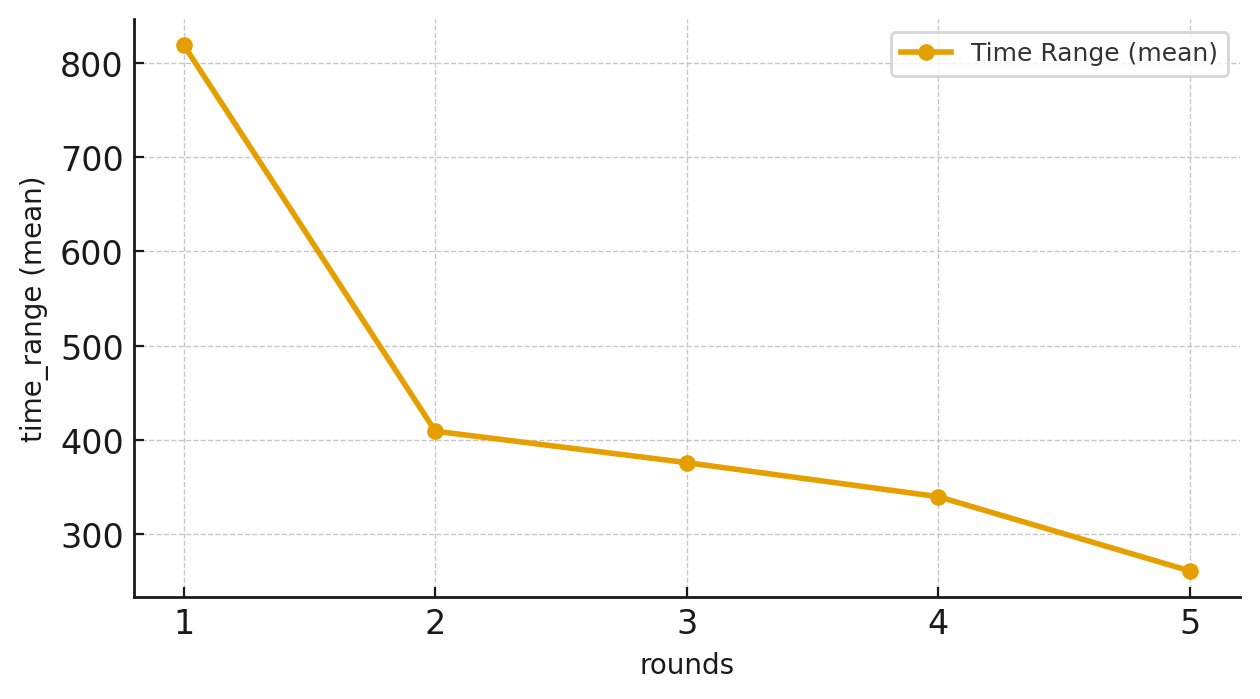}
        \caption{Average time range}
        \label{fig:eva_timerange}
    \end{subfigure}

    \caption{EVA statistics across rounds.}
    \label{fig:eva_stats_all}
\end{figure*}

\paragraph{How the EVA attribute its computation in each round.}
To understand why the EVA schema surpasses both the baseline approaches and the traditional agentic paradigm, we further analyze the distribution of visual-token usage across rounds, which highlights the autonomous behavior of our agent compared with other methods.
We compute the visual-token attribution for EVA and contrast it with the baselines, which allocate all visual tokens in the first round. Figure \ref{fig:eva_stats_all} illustrates the distributions of frames (nframe × resize), nframes, resize, and time range over different rounds.
From these results, we observe that EVA initially explores the video using a large number of frames and a long temporal span. However, in the second round, both nframes and the time range drop sharply, while the resize factor continues to increase, indicating that EVA zooms in to gather more fine-grained information. This progression clearly demonstrates EVA’s agentic decision-making capability.

\subsection{Evaluation on Non-Multi-Choice Benchmark}
ELV-Halluc~\citep{lu2025elvhallucbenchmarkingsemanticaggregation} is a benchmark designed to evaluate semantic aggregation hallucination (SAH) in multimodal large language models (MLLMs) using in-video and out-video QA pairs. The SAH ratio is defined as:
\[
\text{SAH Ratio} = \frac{\text{OutAcc} - \text{InAcc}}{1 - \text{InAcc}}
\]
SAH denotes the phenomenon where a model accurately perceives frame-level semantic information but generates hallucinations during the aggregation of this information into event-level semantics. This unique failure mode necessitates that models first achieve precise frame-level visual perception while demonstrating robust temporal localization capabilities—particularly proficiency in understanding temporal sequences of video events.

Table \ref{tab:main_results} presents the key experimental results on the ELV-Halluc benchmark. Specifically, our proposed EVA model achieves the highest Overall Accuracy among all compared models and reduces the SAH-ratio from 8.8\% to 5\% in contrast to Qwen2.5-VL-7B. Such superior performance fully demonstrates the remarkable capability of EVA: on one hand, our agentic tool-call framework provides the model with more visual details during the required time intervals, which effectively enhances the model’s frame-level perception ability; on the other hand, the reinforcement learning (RL)-based training paradigm requires the model to accurately locate the timestamps that necessitate tool calls during the training process, thereby consolidating the model’s mapping ability between semantics and temporal sequence.
\begin{table*}[ht!]
\centering
\renewcommand{\arraystretch}{1.3}
\resizebox{\textwidth}{!}{%
\begin{tabular}{ccccccccccccccccc}
\hline
\multirow{2}{*}{\textbf{Models}}  &  \multirow{2}{*}{\textbf{LLM size}} &
\multicolumn{3}{c}{\textbf{Visual Details}} &
\multicolumn{3}{c}{\textbf{Object}} &
\multicolumn{3}{c}{\textbf{Action}} &
\multicolumn{3}{c}{\textbf{Declarative Content}} &\multirow{2}{*}{\textbf{Avg Acc↑}}
& \multirow{2}{*}{\textbf{Avg Diff.↓}}&\multirow{2}{*}{\textbf{SAH Ratio↓}}\\
& &In. & Out. & Diff. &
  In. & Out. & Diff. &
  In. & Out. & Diff. &
  In. & Out. & Diff. &
   &  & \\
\hline
\multicolumn{16}{c}{\textbf{Closed Source Models}} \\
\hline
GPT-4o &/& 8.5 & 14.2 & 5.7 & 16.3 & 17.5 & 1.2 & 13 & 15.2 & 2.2 & 13.3 & 12.2 & -1.1 & 13.7 & 2 & 2.2 \\

\hline
\multicolumn{16}{c}{\textbf{Open Source Models}} \\
\hline
Qwen2.5VL-3B &3B& 2.2& 10.5& 8.3& 7.7& 13.8& 6.1& 5& 8& 3& 6& 6& 0& 7.4& 4.3&4.5\\
LLaVA-OV-7b & 7B& 4.5 & 16.5 & 12 & 6.75 & 11.5 & 4.75 & 3.5 & 9.5 & 6 & 6 & 7 & 1 & 8.1  & 5.9 & 6.2 \\
Qwen2.5VL-7b &7B& 10.2 & 26& 15.8& 17.5& 30.7& 13.2& 13& 20.7& 7.7& 16.8& 10.5& -6.3& 18.1&7.6 &8.8\\
InternVL3-8B &7B& 12.5 & 19.5 & 7.0 & 14.5 & 19.5 & 5.0 & 13.5 & 20.5 & 7.0 & 12.8 & 17.7 & 4.9 & 16.3 & 5.9 & 6.8 \\
InternVL3-14B &14B& 17.5 & 24.5 & 7.0 & 22.8 & 24.5 & 1.7 & 16.3 & 17.7 & 1.4 & 15.2 & 15.5 & 0.3 & 19.2 & 2.6 & 3.1 \\
Qwen2.5VL-32B &32B& 16.5 & 24.5 & 8.0 & 21.7 & 24.5 & 2.8 & 17.2 & 15.0 & -2.2 & 15.2 & 7.2 & -8 & 17.7 & 0.1 & 0.2 \\
InternVL3-38B &32B& 25.3 & 29   & 3.7  & 24.2 & 28   & 3.8  & 24   & 30   & 6    & 24.5 & 24.2 & -0.3  & 26.1 & 3.3  & 4.3 \\

\hline
\multicolumn{16}{c}{\textbf{Our Model}} \\
\hline
EVA  & 7B & 21.7& 27.6 & 5.9 & 24.1  & 25.3 & 1.2  & 27.6  & 32.3 & 4.7 & 23.2 & 26.5 & 3.3 & \textbf{26.2} & 3.8  & 5.0 \\
\hline

\rowcolor{gray!20} 

\hline
\end{tabular}
}
\caption{Main results on ELV-Halluc. Diff. denotes the gap between in-video and out-video accuracy. All accuracies are shown as percentages.}
\label{tab:main_results}
\end{table*}

\subsection{Evaluation Details}
We use \texttt{vLLM} to serve EVA. Our frame selection tool extracts frames in \texttt{jpg} format, which are then fed into the next round of processing. All evaluations are conducted using the original video resolution (720p).

For the baselines, we evaluate Qwen2.5-VL with video input. We set \texttt{min\_pixels} to $1280 \times 28 \times 28$ and \texttt{max\_pixels} to $16384 \times 28 \times 28$.

\subsection{Prompts}
We show the prompts for the reflective thinker in the section, revealing how we construct high-quality training data.

\paragraph{Prompt for the Reflective Thinker} Prompt \ref{reflector_prompt} is the prompt for the reflector.
\begin{figure*}
\begin{promptbox}
    You are the REFLECTOR agent that audits the EXECUTOR agent's planning and the current tool plan, and fixes mistakes before the tool is executed.

You will be given:
- Video duration and question
- The EXECUTOR's last round (Round N-1) thoughts including Summary/Analysis/Plan and its tool call JSON, with computed stats (fps and visual budget)
- The newly proposed tool call for Round N, with computed stats

Your job: decide whether the Round N plan violates any of the following rules. If any rule is violated, output a corrected Analysis, Plan, and tool call JSON. Keep the Summary from Round N UNCHANGED (the caller will stitch it back). If there is NO issue, think first then output <NO_CHANGE>.

Rules to check:
1) If the global sampling did NOT find information, do NOT randomly select a video segment (e.g., middle part). Instead, increase the global sampling density and resolution AGGRESSIVELY: raise resize from ~0.1 to at least 0.5; increase nframes by at least 3x, while keeping fps ≤ 1.
2) If the model want to focus on a specific time range according to the visual information, that is correct and just do not change it.
2) If a specific segment was already densely sampled (fps == 1) in Round N-1, do NOT continue dense sampling on the same segment next; instead increase global sampling density/resolution to search globally.
3) If the visual budget in a tool call is too small (< 15), increase it to at least 15 by adjusting resize and/or nframes.
4) If fps > 1, set fps = 1 by reducing nframes to at most end_time - start_time.
5) If there is no problem, after thinking output <NO_CHANGE> as the only content.
\end{promptbox}
\caption{Reflector Prompt}
\label{reflector_prompt}
\end{figure*}

\subsection{Case Visualization}

\begin{figure*}

\begin{dialoguebox}[SYSTEM]{sys-bg}
Use Frame Select Tool to Analyze the video and generate an answer to the question.
\end{dialoguebox}

\begin{dialoguebox}[USER]{sys-bg}
Video Length: 397 seconds. Original video resolution: 720p.\\
What triggers the 'GOAAAAAL!!' animation during gameplay?
\end{dialoguebox}

\begin{dialoguebox}[ASSISTANT]{bot-bg}
{\itshape

To tackle this, I first need to understand what the question is asking: I'm looking for the specific moment that triggers the 'GOAAAAAL!!' animation during gameplay. This means I have to find a frame that clearly shows this animation and identify its cause.

Next, I consider all the potential strategies at my disposal. There’s dense sampling, sparse sampling, keyframe extraction, and hybrid approaches like the one I’m leaning towards—customized sampling combined with aggressive resizing. Each method has its own set of arguments that I can tweak to suit my needs better.

Dense sampling would give me lots of frames across the entire video, which might be useful for spotting the exact moment. However, it could also be overwhelming and computationally expensive. Keyframe extraction aims to cover the essence of the video by focusing on significant frames, but it may not provide enough detail about the gameplay nuances. Sparse sampling might give me an overview of the whole video. 
}

\begin{lstlisting}[language=json,basicstyle=\tiny\ttfamily]
{
  "tool": "frame_select",
  "arguments": {
    "start_time": 0, "end_time": 397, "nframes": 10, "resize": 0.1
  }
}
\end{lstlisting}
\end{dialoguebox}

\begin{dialoguebox}[USER (Tool Response: Round 1)]{sys-bg}
\foreach \i in {0,1,2,3,4,5,6,7,8,9} {
    \includegraphics[width=0.10\linewidth]{figures/case2/round1/frame_000\i.jpg}\hspace{0.5pt}%
}
\\[3pt]
If more information is needed, call the frame selection tool again.
Question: What triggers the 'GOAAAAAL!!' animation during gameplay?
\end{dialoguebox}

\begin{dialoguebox}[ASSISTANT]{bot-bg}
{\itshape

From the low-resolution frames provided, we see various gameplay scenes involving characters and text references like "GOAAAAAL!!". However, the exact trigger for the 'GOAAAAAL!!' animation isn't clear. The scenes with players positioned near goals or scoring actions seem promising leads.

To pinpoint the exact moment the 'GOAAAAAL!!' animation occurs, I will select a segment of the video around frames showing scoring attempts or celebratory moments. A closer inspection of the given frames suggests that the frame at [03:24] shows a character near a goal post, potentially just before scoring. Therefore, focusing on the segment around this frame for a higher-resolution analysis makes sense.

I'll select a segment from [03:20] to [04:10], capturing 100 frames at a higher resolution (resize=0.4) to get more detail without overloading resources. This should help identify any subtle cues leading up to the 'GOAAAAAL!!' animation.

}

\begin{lstlisting}[language=json,basicstyle=\tiny\ttfamily]
{
  "tool": "frame_select",
  "arguments": {
    "start_time": 200, "end_time": 250, "nframes": 100, "resize": 0.4
  }
}
\end{lstlisting}
\end{dialoguebox}

\begin{dialoguebox}[USER (Tool Response: Round 2)]{sys-bg}
\includegraphics[width=0.10\linewidth]{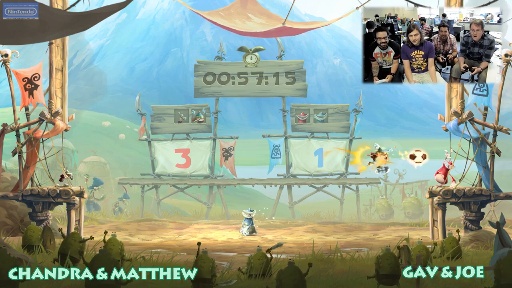}\hspace{0.5pt}%
\foreach \i in {10,20,30,40,50,60,70,80,90} {
    \includegraphics[width=0.10\linewidth]{figures/case2/round2/frame_00\i.jpg}\hspace{0.5pt}%
}
\\[3pt]
\textit{(... Uniformly sampled 10 frames from 100 frames: indices 0, 10, 20, ..., 90 ...)}
\\[3pt]
If more information is needed, call the frame selection tool again.
Question: What triggers the 'GOAAAAAL!!' animation during gameplay?
\end{dialoguebox}

\begin{dialoguebox}[ASSISTANT]{bot-bg}
{\itshape

The increased resolution has provided clearer visuals, particularly highlighting the 'GOAAAAAL!!' animations that coincide with successful goals. Observing the frames, the animation appears when a character scores a goal, as indicated by the ball entering the opposing team's goal area. The transition to the celebration scene, including the activation of the 'GOAAAAAL!!' text and confetti, corresponds directly to the ball crossing the goal line.

Based on the gameplay analyzed so far, the key triggers for the 'GOAAAAAL!!' animation appear to be:

1. The ball reaching the opponent's goal area.
2. The scoring team's character celebrating the goal.

Given these observations, no further analysis is necessary since the pattern indicating how the animation is triggered is evident. The frames demonstrate that the animation is directly linked to scoring actions within the game.

}

\vspace{5pt}
\textbf{Answer:} The 'GOAAAAAL!!' animation is triggered when the ball successfully enters the opponent's goal area, signaling a successful score in the game. The frames show that the celebration and animation activate immediately after the ball crosses the goal line, confirming this as the primary trigger.
\end{dialoguebox}
\caption{Three-Round Case}
\label{fig:case1}
\end{figure*}

\paragraph{Success Case: Multi-turn Grounding and Zoom-In}
Figure \ref{fig:case1} shows a success case. Given the query and the video length, the EVA carefully ponders several potential actions, including dense sampling, sparse sampling, and keyframe extraction. It then thinks about what new information each action will bring and what they will cost. Finally, it generates an efficient tool-call to get a brief overview of the video. In the second round, the EVA generates a new tool call based on the first-round frames, allocating enough visual budget to the corresponding segment.

\paragraph{Success Case: Direct Dense Sampling}
Given the query, the EVA autonomously plans and figures out the most efficient way to find the correct answer. Some Video-QA questions do need dense and high-resolution frames, just like what baseline passive MLLMs do. In these cases, the EVA will go back to the traditional unified sampling method and generate the correct answer after the first tool call. Figure \ref{fig:case2} shows an example.

\begin{figure*}

\begin{dialoguebox}[SYSTEM]{sys-bg}
Use Frame Select Tool to Analyze the video and generate an answer to the question.
\end{dialoguebox}

\begin{dialoguebox}[USER]{sys-bg}
Video Length: 1218 seconds. Original video resolution: 384p.\\
What was the process followed by the camera wearer to set up the sewing machine during the holiday-themed Scene 6?\\
A: Set up the sewing station, placed tools on the table, connected the machine to power, threaded the needle, and began sewing.\\
B: Positioned sewing materials, removed the cover, connected power cables, wound the bobbin, threaded the machine, and powered on.\\
C: Cleared the workspace, uncovered the sewing machine, connected necessary cables, adjusted the needle, and wound the bobbin.\\
D: Moved to the table, arranged sewing items, uncovered the machine, connected cables, powered on, prepared and threaded the machine.
\end{dialoguebox}

\begin{dialoguebox}[ASSISTANT]{bot-bg}
{\itshape
To answer the user’s question about the process followed by the camera wearer to set up the sewing machine during the holiday-themed Scene 6, I need to analyze the video content. Since the video length is 1218 seconds, I'll start by sampling frames from the entire video to get an overview. This should help me identify the relevant scene and the process described in the options.
}
\begin{lstlisting}[language=json,basicstyle=\tiny\ttfamily]
{
  "tool": "frame_select",
  "arguments": {
    "start_time": 0,
    "end_time": 1218,
    "nframes": 60,
    "resize": 0.5
  }
}
\end{lstlisting}
\end{dialoguebox}

\begin{dialoguebox}[USER (Tool Response)]{sys-bg}
\foreach \i in {00,05,10,15,20,25,30,35,40,45} {
    \includegraphics[width=0.19\linewidth]{figures/case3/round1/frame_00\i.jpg}\hspace{0.5pt}%
}

\textit{(... Displaying first 10 of 60 frames ...)}

If more information is needed, call the frame selection tool again.\\
Question: What was the process followed by the camera wearer to set up the sewing machine during the holiday-themed Scene 6?\\
A: Set up the sewing station, placed tools on the table, connected the machine to power, threaded the needle, and began sewing.\\
B: Positioned sewing materials, removed the cover, connected power cables, wound the bobbin, threaded the machine, and powered on.\\
C: Cleared the workspace, uncovered the sewing machine, connected necessary cables, adjusted the needle, and wound the bobbin.\\
D: Moved to the table, arranged sewing items, uncovered the machine, connected cables, powered on, prepared and threaded the machine.
\end{dialoguebox}

\begin{dialoguebox}[ASSISTANT]{bot-bg}
{\itshape
The video shows the person setting up the sewing machine by connecting cables, removing the cover, and threading the machine. The process aligns with option D.
}
\vspace{5pt}
\textbf{Answer:} D
\end{dialoguebox}
\caption{Two-Round Case}
\label{fig:case2}
\end{figure*}
\end{document}